\def\year{2020}\relax
\documentclass[letterpaper]{article} % DO NOT CHANGE THIS
\usepackage{aaai20}  % DO NOT CHANGE THIS
\usepackage{times}  % DO NOT CHANGE THIS
\usepackage{helvet} % DO NOT CHANGE THIS
\usepackage{courier}  % DO NOT CHANGE THIS
\usepackage[hyphens]{url}  % DO NOT CHANGE THIS
\usepackage{graphicx} % DO NOT CHANGE THIS
\usepackage{graphicx}  % DO NOT CHANGE THIS
\usepackage{paralist, tabularx}
\usepackage{multirow}
\usepackage{caption}
\usepackage{comment}
\usepackage{amssymb}
\usepackage{amsmath}
\usepackage{float}

\newcommand\Vtextvisiblespace[1][.3em]{%
	\mbox{\kern.06em\vrule height.3ex}%
	\vbox{\hrule width#1}%
	\hbox{\vrule height.3ex}}

\usepackage{algorithmic}
\usepackage[ruled,vlined]{algorithm2e}
\usepackage{booktabs}

\usepackage{CJKutf8}
\usepackage{CJK}
\usepackage{url}
\usepackage[dvipsnames]{xcolor}

\title{Mastering Complex Control in MOBA Games with Deep Reinforcement Learning}
\author{Deheng Ye \textsuperscript{\rm 1}, Zhao Liu \textsuperscript{\rm 1}, Mingfei Sun \textsuperscript{\rm 1} \thanks{Work done as research interns at Tencent.}, Bei Shi \textsuperscript{\rm 1}, Peilin Zhao \textsuperscript{\rm 1}, Hao Wu \textsuperscript{\rm 1} \footnotemark[1], Hongsheng Yu \textsuperscript{\rm 1}, \vspace{0.5mm}
\\ \vspace{0.5mm} \Large \textbf{Shaojie Yang \textsuperscript{\rm 1}, Xipeng Wu \textsuperscript{\rm 1}, Qingwei Guo \textsuperscript{\rm 1}, Qiaobo Chen \textsuperscript{\rm 1}, Yinyuting Yin \textsuperscript{\rm 1}, Hao Zhang \textsuperscript{\rm 1},}
\\ \vspace{1mm} \Large \textbf{Tengfei Shi \textsuperscript{\rm 1}, Liang Wang \textsuperscript{\rm 1}, Qiang Fu \textsuperscript{\rm 1}, Wei Yang \textsuperscript{\rm 1}, Lanxiao Huang \textsuperscript{\rm 2} }\\ 
\textsuperscript{\rm 1} Tencent AI Lab, Shenzhen, China\\ 
\textsuperscript{\rm 2} Tencent Timi Studio, Chengdu, China \\ 
\{dericye,ricardoliu,mingfeisun,beishi,masonzhao,alberthwu,yannickyu,shaojieyang,haroldwu,leoqwguo,\\ciaochen,mailyyin,howezhang,francisshi,enginewang,leonfu,willyang,jackiehuang\}@tencent.com % email address must be in roman text type, not monospace or sans serif
}

\begin{document}

\newcommand{\red}[1]{\textcolor{red}{#1}}
\newcommand{\blue}[1]{\textcolor{blue}{#1}}

% make the title area
\maketitle

\begin{abstract}
We study the reinforcement learning problem of complex action control in the Multi-player Online Battle Arena (MOBA) 1v1 games. 
This problem involves 
far more complicated state and action spaces than those of traditional 1v1 games, such as Go and Atari series, which makes it very difficult to search any policies with human-level performance. 
In this paper, we present a deep reinforcement learning framework to tackle this problem from the perspectives of both system and algorithm. 
Our system is of low coupling and high scalability, which enables efficient explorations at large scale.
Our algorithm includes several novel strategies, including control dependency decoupling, action mask, target attention, and dual-clip PPO, with which our proposed actor-critic network can be effectively trained in our system. 
Tested on the MOBA game \textit{Honor of Kings}, our AI agent, called \textit{Tencent Solo}, can defeat top professional human players in full 1v1 games.
\end{abstract}

\section{Introduction} \label{sec:intro}
% introducing the problem and game
Deep reinforcement learning~(DRL) has been widely used for building agents to learn complex control in competitive environments. 
%In the 
In the competitive setting, a considerable amount of existing DRL research adopt two-agent games as the testbed, i.e., one agent versus another (1v1). 
Among them, Atari series and board games have been widely studied.  %\cite{mnih2015human,silver2016mastering}. 
%With DRL, an agent can usually learn the hidden representation of the environment as well as the common knowledge to handle various game situations. 
For example, a human-level agent for playing Atari games is trained with deep Q-networks~\cite{mnih2015human}. The incorporation of supervised learning and self-play into the training brings the agent to the level of beating human professionals in the game of Go~\cite{silver2016mastering}.
And recently, a more general DRL method is further applied to the Chess and Shogi 1v1 games \cite{silver2017mastering}. 
%Such success of reinforcement learning in the aforementioned games further leads us to one question: is it possible (and how) to use deep reinforcement learning to train an agent which can master even more complex controls. To investigate this question, 

In this paper, we move on to one type of 1v1 games that has the next level of complexity, i.e., the MOBA 1v1 games. 
As we know, RTS games are considered as a grand challenge for AI research \cite{alphastarblog,silva2017moba}. 
MOBA 1v1 is a real-time strategy (RTS) game that requires highly complex action control. 
%, in which agents must learn to plan, select target units, control skill combos, induce, and deceive the opponents. 
%And the correlation between player skill and actions-per-minute is not strong. 
Compared with traditional 1v1 games, e.g., board and Atari, MOBA 1v1 games have far more complicated environments and controls. % than that of the Atari and Go games. 
Take the MOBA 1v1 games in \textit{Honor of Kings} as an example, the magnitude of states and actions involved can reach to $10^{600}$ and $10^{18000}$, while these in Go are $10^{170}$ and $10^{360}$ \cite{silver2016mastering}, illustrated in Table~\ref{table:complexity}. 

%MOBA 1v1 games have three characteristics that distinguish its complexity level from others. 
%First, it is a Real-Time Strategy (RTS) game, where the agent needs to make instant controls based on the current environment dynamics. 
% Specifically, at each timestep, the agent has only a local view of its environment, while in the Atari and Go games the agent has the global view. The lack of full knowledge in MOBA games hinders the agent's learning of the effective representations of the environment, and potentially increases the difficulty of searching the real-time optimal strategy search, which is the key to win the game.  
%\textcolor{red}{With this complexity, simply applying deep reinforcement learning algorithms in Go and Atari to MOBA games is not working any longer. }
Besides, the complexity of MOBA 1v1 also comes from the playing mechanism. 
%And the correlation between player skill and actions-per-minute is not strong.
%The state of the environment is partially observable. 
To win a game, in the partially observable environment, agents must learn to plan, attack, defend, control skill combos, induce, and deceive the opponents. 
%The correlation between player skill and actions-per-minute is not strong.
Apart from the player's and the opponent's agent, there exists many more game units, e.g., creeps and turrets. 
This creates challenge to the target selection which requires %on which enemy unit to attack or which own unit to protect.
delicate sequences of decision making and corresponding action controls. 
Furthermore, different heroes in a MOBA game have very different playing methods. The action control can completely change from hero to hero, which calls for robust and unified modeling. %This differs from traditional 1v1 games. 
Last but not least, there lacks high-quality human game data for MOBA 1v1 which makes supervised learning unfeasible, because players generally use the 1v1 mode to practice heroes, while the MOBA 5v5 mode is used for formal matches in mainstream MOBA games, like \textit{Dota} and \textit{Honor of Kings}. 
Note that in this paper we focus on MOBA 1v1 games rather than MOBA 5v5 as the latter emphasizes more on the team collaborative strategy of all agents than the action control of any single agent. In this regard, the MOBA 1v1 setting is more appropriate to study the problem of complex control in games.

\begin{table}[t!]
\scriptsize
\centering
\caption{Comparing Go and MOBA 1v1}
%\vspace{-0.1in}
\label{table:complexity}
\begin{tabular}{@{}c@{}@{}c@{}@{}c@{}}
\toprule
Game & \textbf{Go 1v1} & \textbf{MOBA 1v1} \\ \midrule
Action space & \begin{tabular}[c]{@{}c@{}}$250^{150}\approx10^{360}$ \\ (250 pos available, 150 \\ decisions per game on average)\end{tabular} & \begin{tabular}[c]{@{}c@{}}$10^{18000}$ (100+ discretized actions, \\ 9,000 frames per game)\end{tabular} \\ \midrule
State space & \begin{tabular}[c]{@{}c@{}}$3^{361}\approx10^{170}$ \\(361 pos, 3 states each)\end{tabular} & \begin{tabular}[c]{@{}c@{}}$2^{2000}\approx10^{600}$ (2 heroes, \\ (1000+ pos)*(2+ states))\end{tabular} \\ \midrule
Human player data  & rich, high-quality & little\\  \midrule
Peculiarity &  long-term tactics & real-time, complex control\\  \bottomrule
%Typical method & Supervised+Reinforcement & NA\\  \bottomrule
\end{tabular}
%\vspace{-5mm}
\normalsize
\end{table}

To handle these challenges, 
we design a deep reinforcement learning framework, together with a set of algorithm-level innovations, to enable efficient explorations at massive scale for multi-agent competitive environments like MOBA 1v1 games. 
%that enables massive scale of explorations multi-agent competitive environments like MOBA 1v1 games. 
% Our framework can cope with the complex control of agents using efficient sample generation and distributed training.
We design a neural network architecture including the encoding of multi-modal inputs, the decoupling of inter-correlations in controls, exploration pruning mechanism, and attack attention, to consider the ever-changing game situations in MOBA 1v1 games. 
%In the experiment, following the evaluation methods proposed by \cite{jaderberg2019human}, we held the tournament evaluation to assess the generalization performance of our agents during the training process and different versions of our agents. 
To evaluate the upper limit and the robustness of the trained AI agents thoroughly, we invite professional players and a variety of top-amateur human players to compete with our AI agents. 
We also compare our method with existing state-of-the-art works on building MOBA 1v1 agents \cite{jiang2018feedback}. 
%Our AI has mastered very complicated behaviors. The matches between AI and professional players are highlighted here: \url{https://dwz.cn/F8wT4nBB}. 
%, which demonstrates the superhuman performance of our framework. 
Our contributions are as follows:
\begin{itemize}
    \item We present a systematic and thorough study on building AI for playing MOBA 1v1 games, which require highly complex action control of agents. 
        On system aspect, we develop a deep reinforcement learning framework which provides scalable and off-policy training. On algorithm aspect, we develop an actor-critic neural network for modeling MOBA action controls. Our network optimizes with a multi-label proximal policy algorithm (PPO) objective, and is featured with the decoupling of control dependency, an attention mechanism for target selection, action mask for efficient exploration, LSTM for learning skill combos, and an improved version of PPO, called dual-clip PPO, for ensured training convergence. 
    %We design a novel network architecture which can handle various game situations such as varied number of units, uncertain target units and action masks.
    \item Extensive experiments show that the trained AI agent, called \textit{Tencent Solo}, can defeat top professional human players on different hero types, tested on the 1v1 mode of \textit{Honor of Kings}, a popular MOBA game. 
    %In the future, we will publish our MOBA environment and SUPEX framework as the standard testbed and baseline method for reinforcement learning research community. 
\end{itemize}

\section{Preliminaries} \label{sec:related}

%In this section, we review related work on the agent control and reinforcement learning domain, provide necessary notations, and some background on MOBA 1v1 games. 

\subsection{Notation}
We focus on the two-agent world for multi-agent Markov games~\cite{bansal2017emergent}, which can be extended to multiple agents.  
We use the tuple $(\mathcal{S},  \mathcal{O}, \mathcal{A}, P, r, \rho_0, \gamma)$ to denote an infinite-horizon, discounted Markov Decision Process, where $\mathcal{S}$ is the state space, $\mathcal{O}$ is the observable state space of each agent, $\mathcal{A}$ is the action space, 
$P:\mathcal{S}\times\mathcal{A}\rightarrow\mathcal{S}$ denotes the state transition probability, 
%$P:\mathcal{S}\times\mathcal{A}\times\mathcal{S}\rightarrow \mathbb{R}_{+}$ denotes the state transition probability, 
$r:\mathcal{S}\times\mathcal{A}\rightarrow\mathbb{R}$ represents the reward function, $\rho_0:\mathcal{S} \rightarrow\mathbb{R}$ is the distribution of the initial state $s_0$, and $\gamma\in(0, 1]$ is the discount factor. 
A stochastic policy $\pi$ %in the policy set $\Pi$ 
is a mapping $\mathcal{O}\times\mathcal{A}\rightarrow [0, 1]$. 
%Let $\tau$ denote a trajectory sampled from $\pi$: $\tau = \big[ (s_0, a_0), (s_1, a_1), ..., (s_n, a_n) \big]$. 
In our complex control problem, each agent aims to maximize the cumulative reward returns, i.e., the objective $\mathbb{E}\big[\sum_{t=0}^{T}\gamma^{t}r(s_t, a_t)\big]$, where $T$ is the time horizon. 
%i.e., the objective $\mathbb{E}_{\pi}\big[ r(s, a) \big] \triangleq \mathbb{E}\big[\sum_{t=0}^{T}\gamma^{t}r(s_t, a_t)\big]$, where $T$ is the time horizon. 
%we use the expectation with respect to a policy $\pi$ to denote the expectation with respect to the trajectories it generates: $\mathbb{E}_{\pi}\big[ r(s, a) \big]\triangleq \mathbb{E}\big[\sum_{t=0}^{\infty}\gamma^{t}r(s_t, a_t)\big]$, where $s_0\sim\rho_0$, $a_t\sim\pi(a_t|s_t)$, $s_{t+1}\sim P(s_{t+1}|a_t, s_t)$. 
%In reinforcement learning, the task is to maximize the cumulative returns, i.e., to maximize the objective $\mathbb{E}_{\pi}\big[ r(s, a) \big]$.

\subsection{Related Work}
Multi-agent control with reinforcement learning has two settings: cooperative setting and competitive setting. Our work belongs to the latter. 
A large proportion of existing works use games as the testbed for RL advances.

For the cooperative setting, a survey is done by \cite{panait2005cooperative}. 
Recently, Foerster et al. \shortcite{foerster2016learning} study multi-agent cooperation to solve riddles with recurrent Q-networks. 
Collaborative agents for playing 3D FPS games have also been explored \cite{jaderberg2019human,lample2017playing}.
Another recent work builds a macro-strategy model for guiding multi-agents in MOBA 5v5 games using supervised learning \cite{wu2019hierarchical}. 
%aiming to provide a cooperative and unified guidance for multi-agents. %, and is based supervised learning on user game data. 
%They divide a whole MOBA game into several game phases, and train a model to guide which region in the map is suggested go during each phase. However, they only focus on the modeling of macro-strategy, i.e., providing a rough guidance on where to go in the game map, rather than micro-management, which can tell what specific actions a hero should take.  Thus, their method can not be used separately as a game AI solution. By comparison, we directly model the action control in MOBA. 

For the competitive setting, 1v1 games are heavily studied. 
%Some studies: \cite{campbell2002deep,silver2016mastering,mnih2015human,lample2017playing,ontanon2013survey,silva2017moba}. 
A typical work is AlphaGo \cite{silver2016mastering}, which combines supervised learning and RL. 
RL has also been successfully applied to Atari games \cite{mnih2015human}, which contain both single-agent games, and multi-agent games like the 1v1 Pong game. 
Further, He et al. \shortcite{he2016opponent} focus on opponent modeling in competitive setting via deep Q-learning, using a simulated soccer game and a trivia game as testbed. 
Tampuu et al. \shortcite{tampuu2017multiagent} use deep Q-learning to train the 1v1 Pong game agents. 
Bansal et al. \shortcite{bansal2017emergent} construct four 1v1 games in MuJoCo environment to analyze the emergent complexity in multi-agent competition.

Comparing to these, we study RL systems with a more complex competitive setting, i.e., the MOBA 1v1 games. 
One published state-of-the-art work on this line proposes a monte-carlo tree search (MCTS) based RL method for playing MOBA 1v1, which uses the game \textit{Honor of Kings} as testbed~\cite{jiang2018feedback}. 
Recently, OpenAI announced an AI for playing Dota 2 (a popular MOBA game) that can defeat professionals. 
Their technical details are not open yet \footnote{by Nov. 10, 2019, the notification date of AAAI-2020. }, an overview is posted \cite{OpenAI_dota}. 
Apart from this, our work presents the first systematic investigation focusing on the action controls of agents in complex games (known as micro-management in esports). 
We play 1v1 full game in \textit{Honor of Kings}, i.e., the game ends until the home base destroyed; %, while OpenAI plays with an earlier termination condition; 
And further, the robustness and scalability of our method have been thoroughly tested, as it has been applied to train various hero types, e.g., Mage, Marksman, Warrior, etc. 
As mentioned, playing a different MOBA hero is like playing a different game. 
%, while OpenAI only plays one hero of one type, i.e., Shadow Fiend. 
%3) we develop efficient pruning and randomization for fast training. 

This work is also related to research on building AI agents for playing StarCraft 1v1 games which have been significantly explored \cite{ontanon2013survey,robertson2014review}. 
By comparison, StarCraft 1v1 games are of a different kind of complexity, i.e., 
%the player has to control many more game units than that of MOBA, but the action control mechanism of each individual unit is simpler. Therefore, 
MOBA 1v1 is known for the complex action control of heroes which is in the scope of this paper, while StarCraft 1v1 measures more on the strategy to control many game units simultaneously. 

%\input {sections/04-1-framework.tex}

%\begin{figure*}
%    \centering
%    \includegraphics[width=1.0\linewidth]{images/overview_v2.png}
%    \caption{Overview of our SUPEX framework.}
%    \label{fig:overview}
% \end{figure*}

\section{System Design} \label{sec:framework}
% This section is about the method. 
% We can talk about implementation of reinforcement learning system, and the advantage of policy and value function, such as feature engineering, reward engineering, attention strategy, LSTM implementation.

% Besides, in order to bootstrap the robust of our framework, we take use of  exploration metrics, history policy bootstrapping, opponent sampling.
%Considering complex action control in MOBA games, 
%In this section, we introduce 

In this section, we first present an overview of our proposed framework for MOBA 1v1, and then describe each module. 
To make our presentation easier, we first introduce our system design in this section, and leave  the algorithm design for the next section.

% system scalability + algorithm adaptions (ppo adaption, hierarchical multi-label loss). 

\begin{figure}
    \centering
    \includegraphics[width=\linewidth]{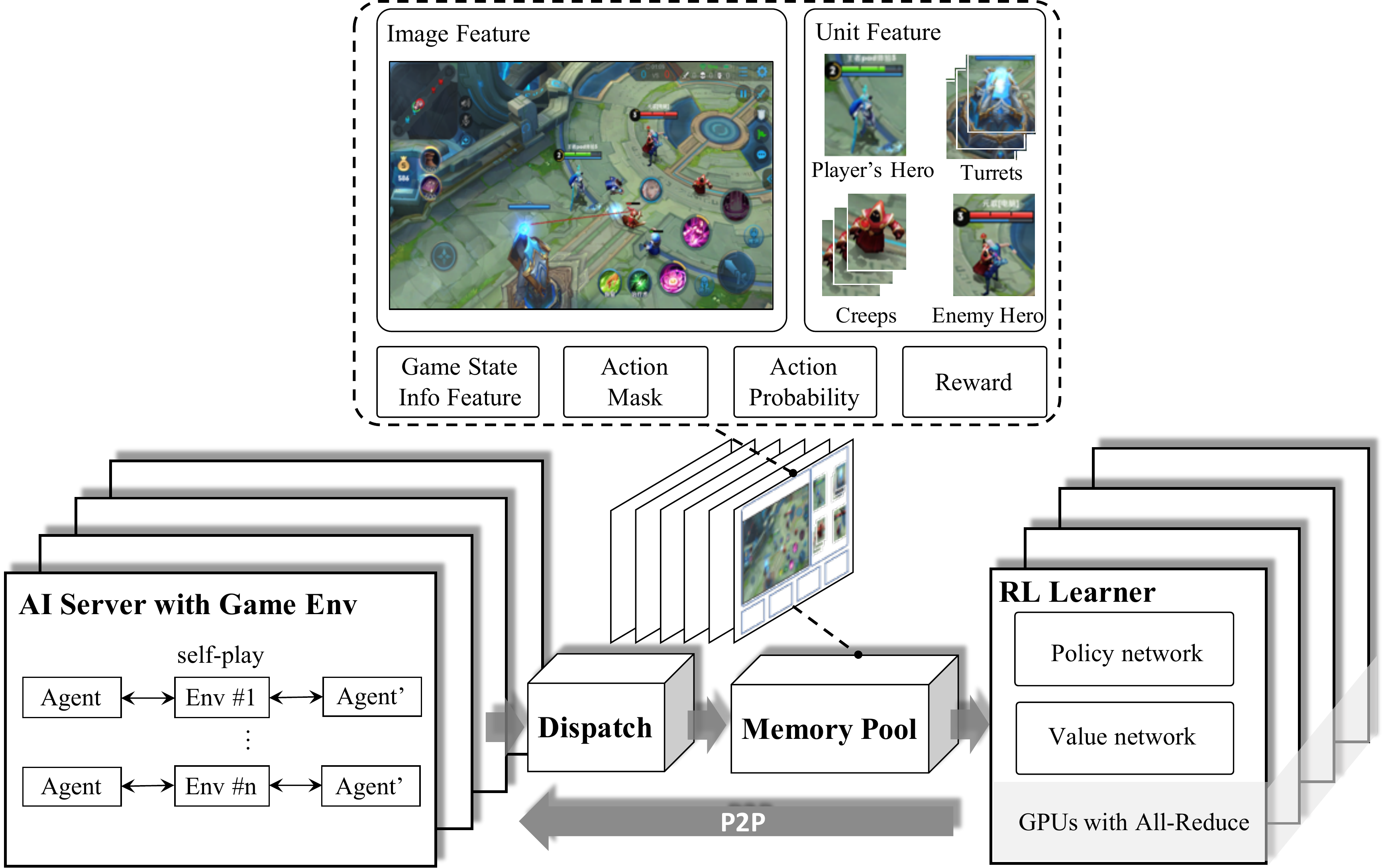}     
    \vspace{-3mm}
    \caption{Overview of our System Design}
   % \vspace{-5mm}
    \label{fig:overview}
\end{figure}

Considering the fact that complex agent control problems can introduce high variance of stochastic gradients, e.g. the MOBA 1v1 games, large batch size is necessary to speed up the training~\cite{mccandlish2018empirical}. 
%Because the MOBA 1v1 games generally require large batch size for training~\cite{mccandlish2018empirical} to reduce the high variance of stochastic gradients,
Thus, we design a scalable and loosely-coupled system architecture to construct the utility of data parallelism.
Specifically, our architecture consists of four modules, i.e., \textbf{Reinforcement Learning (RL) Learner}, \textbf{Artificial Intelligence (AI) Server}, \textbf{Dispatch Module} and \textbf{Memory Pool}, as shown in Fig.~\ref{fig:overview}. 
AI Server implements how the AI model interacts with the environment. 
%Each AI Server is configured to play MOBA 1v1 games with two mirrored policies, i.e., self-play \cite{silver2017mastering}. 
The Dispatch Module is a station for sample collection, compression and transmission.
The Memory Pool is the data storage module, which provides training instances for the RL Learner. 
%The distributed RL Learner acquires samples from memory pool for model training.
%It updates the parameters of policy and value network with many GPUs. 
%After several optimization steps, the updated parameters are synchronized to AI Servers to generate the episodes by the recent policies continuously.
Note that these modules are decoupled and can be flexibly configured, so that our researchers can focus on the algorithm design and the logic of the environment. 
Such a system design is also applicable to other multi-agent competitive problems.
The details of these modules are provided as follows:
%We will make our framework open source. 
%During self-play, feature, reward, action label and action prob are extracted from game episodes at the beginning to decrease the cost of communications and then forwarded by our Dispatch Module to distributed memory pools. 
%After sampling from memory pools, the distributed RL Learners can update the parameters of policy network and value network with multiple GPUs according to the off-policy PPO algorithm~\cite{schulman2017proximal}. After several optimization steps, the updated parameters are synchronized to AI Servers to generate the episodes by the recent policies continuously.
%Note that these modules are decoupled and can be flexibly configured. 
%One only have to change the self-play environment and feature extraction to the studied problem. 
%
%In order to solve RL problems effectively and precisely, such as the King of Honor, our system is implemented as a general-purpose RL training system with several well-designed modules. Our framework consists of four modules: Reinforcement Leanring (RL) Learner, Artificial Intelligence (AI) Server, Game Environment, Dispatch Module and Memory Pool as illustrated in Fig.~\ref{fig:overview}.
% In order to solve RL problems effectively and precisely, such as the King of Honor, our system is implemented as a general-purpose RL training system with several well-designed modules, shown in Fig \ref{fig:system}.

\textbf{AI Server} covers the interaction logic between game environment and the AI model. 
%It is responsible for the model's online inference, feature extraction, reward calculation, value estimation, and historical policy maintenance. 
AI server generates episodes via self-play with mirrored policies \cite{silver2017mastering}. 
%Specifically, the opponent policies are sampled from 70\% of the latest policies and 30\% of the historical policies, which is similar to the treatments in 
The opponent policy sampling is similar to \cite{bansal2017emergent}. 
Based on the features extracted from game state, hero action is predicted using Boltzman exploration \cite{cesa2017boltzmann}, i.e., sampling based on softmax distribution. 
The sampled action is then forwarded to the game core for execution. 
%The predicted action is using the softmax
After execution, the game core returns the corresponding reward value and the next state continuously. 
In use, one AI Server will bind one CPU core. 
Because the game logic deduction runs on CPUs, we also run the model inference on CPUs to save the IO cost. 
In order to generate episodes efficiently, we build a CPU version of the fast inference library FeatherCNN~\footnote{FeatherCNN is a state-of-the-art inference engine for mobile devices: \url{https://github.com/Tencent/FeatherCNN}}. 
FeatherCNN can automatically convert AI models trained from mainstream tools like Tensorflow and Caffe, to a customized format for inference. 

%Considering the performance of inference, we convert the format of our model from Tensorflow into a CPU version of FeatherCNN\footnote{https://github.com/Tencent/FeatherCNN} which increases the speed of generating samples.

%\paragraph{Dispatch Module and Memory Pool} 
%Designed for sample collection, compression and transmission, the 
%Dispatch Module serves as the mediator between the AI Server and the Memory Pool.
Each \textbf{Dispatch Module} is bounded with several AI Servers on the same machine. 
It is a server that collects data samples from AI Servers, consisting of reward, feature, action probabilities, etc. 
These samples are firstly compressed and packed, and then send to Memory Pools. 
The \textbf{Memory Pool} is also a server. Its internals are implemented as a memory efficient circular queue for data storage. 
It supports samples of varied lengths, and data sampling based on the generated time. 
%To save communication, Memory Pool is deployed on the GPU machine with RL Learner to provide training data. 
%pairs of states and actions and then sends them to memory pools with low decay and high concurrency. 
%The memory pool module provides the utility of communications between RL learners and AI Servers. 
%Considering the communication costs, memory pools receive samples via shared memory instead of socket with about 2-3 times of speed boosting.

%\paragraph{RL Learner} 
The \textbf{RL Learner} is a distributed training environment. 
%In order to update parameters of policy network efficiently and implement large batch size training, 
To accelerate policy update using large batch sizes,  multiple RL Learners are integrated to parallelly fetch data from the same number of Memory Pools. 
%Considering communication efficiency between GPUs, 
The gradients in the RL learners are averaged through the ring allreduce algorithm~\cite{sergeev2018horovod}. 
%As mentioned, Memory Pool 
To reduce IO cost, RL Learners communicate with  Memory Pools using shared memory instead of socket, which can deliver 2-3 times of speed boosting. 
%Specifically, the RL Learner launches a thread to open a lock-free queue for the GPU memory to get data from the shared memory. 
%Each RL learner is associated with a same well-designed neural network shown in Figure~\ref{fig:network_arch} and calculates the corresponding gradients with respect to samples from Memory Pools simultaneously.
The trained models from the RL Learners are rapidly synchronized to AI Servers in a peer-to-peer manner.

%\paragraph{Scalability} 
In our system, the experiences generation  is decoupled from  the parameters learning. This flexible mechanism makes AI Servers and RL learners scalable with high throughput. 
%In IMPALA~\cite{espeholt2018impala}, parameters are distributed across the learners, and actors retrieve the parameters from all the learners in parallel. 
%Different from IMPALA, our trained models are synchronized to AI Servers via peer-to-peer from our master RL Learner.
To avoid the bottleneck of communication cost between learners and actors, our trained models are synchronized to AI Servers via peer-to-peer from our master RL Learner. 
To smooth data storage and transmission, we design two mediators, i.e., the Dispatch Server and the Memory Pool Server. 
In practice, we can scale to millions of CPU cores and thousands of GPUs effortlessly. 
Note that such design differs from existing system designs like IMPALA \cite{espeholt2018impala}.
In IMPALA, parameters are distributed across the learners, and actors retrieve the parameters from all the learners in parallel. 

\section{Algorithm Design}

\begin{figure*}
    \centering
    \includegraphics[width=0.85\linewidth]{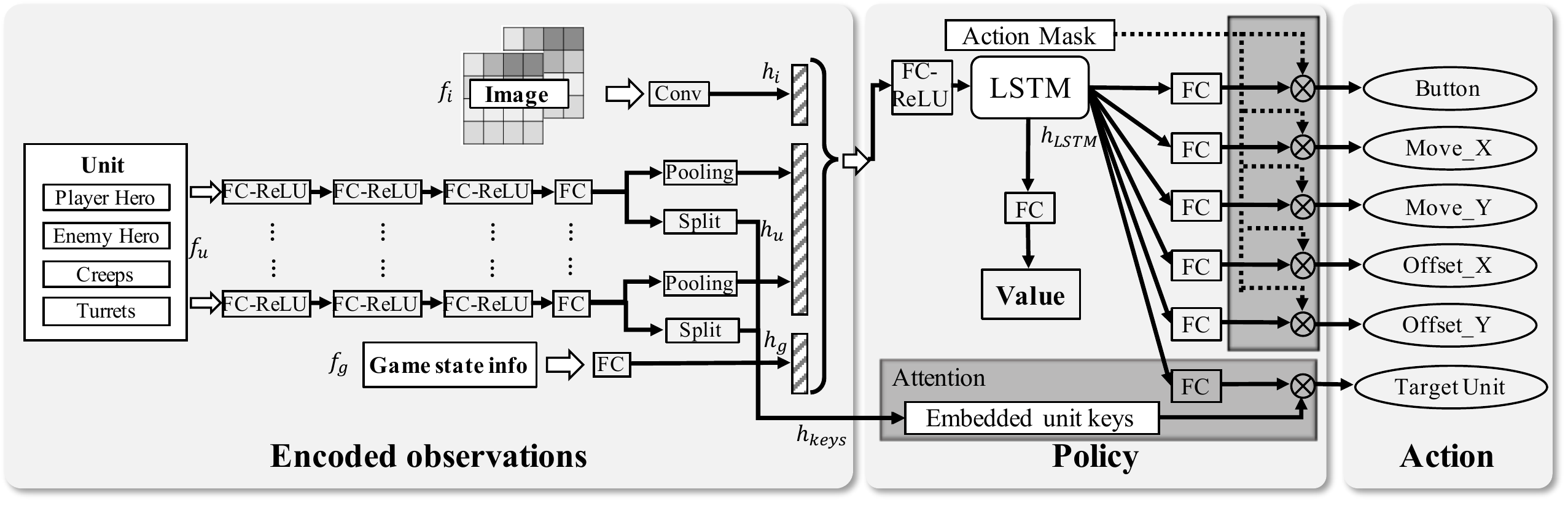}
    \vspace{-0.1in}
    \caption{Illustrations of state, action, policy and value. A state $s\in\mathcal{S}$ covers three types of information: local image info (e.g., obstacles in 2D), observable unit attributes (e.g., hero type, health point), and observable game state info (e.g., game time, turrets destroyed, etc.), i.e., $s = [f_i, f_u, f_g]$. An action $a\in\mathcal{A}$ in a MOBA 1v1 game specifies two items: the content (i.e., the action button to press, the Move\_X, the Move\_Y, Offset\_X, and Offset\_Y) and the target game unit (the action space design is similar to that of OpenAI Five \cite{OpenAI_dota},  which we believe is a general design across MOBA games).  The action buttons include move, attack, skill releasing, etc. The policy $\pi_{\theta}$ is modeled by FCs and an LSTM, which also predicts the values.}
    \label{fig:network_arch}
    \vspace{-0.1in}
\end{figure*}

In the RL Learner, an actor-critic network is implemented to model the action control dependencies in MOBA 1v1 games.
Figure~\ref{fig:network_arch} illustrates this network,
the state and actions. 
To train this network efficiently and effectively, several novel strategies are proposed. 
First, the \textit{target attention mechanism} is designed in this network to help with the target selection in MOBA combats.
Second, LSTMs are leveraged for the hero to learn the skill combos which are critical to create severe and instant damage. 
Third, the decoupling of control dependencies is conducted to form a multi-label proximal policy optimization (PPO) objective. 
Forth, a game-knowledge-based pruning method, called \textit{action mask}, is developed to guide explorations during the reinforcement process.
Finally, a dual-clipped version of the PPO algorithm is proposed to guarantee convergence with large and deviated batches \cite{schulman2017proximal}. 
The details of our network are provided in the remaining paragraphs.

First, the network encodes image features $f_i$, vector features $f_u$, and game state information $f_g$ (the observable game states) as encodings $h_i$,  $h_u$, and $h_{g}$ using convolutions, fully-connections, and fully-connections (FC), respectively. 
%Each group shares parameter. 
Specifically, after a few layers of FC/ReLu, the encoding of $f_u$ is splitted into two parts: the representation of the unit and the attention keys of our target. 
%Note that we perform feature grouping and parameter sharing based on the categorization of the game unit. 
To handle the varied number of units, the same type units are mapped to a feature vector of fixed length by max-pooling. Then the concatenation of $h_i$, $h_u$ for all types, and $h_g$ is represented as the encoding vector of an observable game state.
The state encoding is then mapped to the final representation $h_{LSTM}$ by a LSTM cell, which further takes the temporal information into consideration. 
$h_{LSTM}$ is sent to a FC layer to predict the action $a$.
%\textcolor{red}{seems hard to seperate these components. }
%\paragraph{Target attention} 
The target unit $t$ of action $a$ is predicted by a \textbf{target attention} mechanism over every unit.
This mechanism treats a FC output of $h_{LSTM}$ as the query, the stack of all unit encodings as the keys $h_{keys}$, and calculate the target attention as:
\begin{equation}
\footnotesize
p(t|a) = Softmax(FC(h_{LSTM}) \cdot h_{keys}^T)
\end{equation}
where $p(t|a)$ is the attention distribution over units and the dimension of $p(t|a)$ is the number of units in the state.

%\paragraph{Action mask method}

Second, it is very hard to explicitly model the inter-correlations among different labels in one action of MOBA games in the multi-label policy network, e.g. the correlation between the direction of a skill (Offset\_X and Offset\_Y), and the skill type (Button). To solve this issue, we treat each label in an action independently to decouple their inter-correlations, i.e., \textbf{the decoupling of control dependencies}. 
Before decoupling the inter-correlations, the PPO objective without clipping is:
\begin{equation}
\footnotesize
\max_{\theta} \hat{\mathbb{E}}_{s, a\sim \pi_{\theta_{\text{old}}}}\bigg[ \frac{\pi_{\theta}(a_t|s_t)}{\pi_{\theta_{\text{old}}}(a_t|s_t)} \hat{A}_t \bigg],
\end{equation}
where the expectation $\hat{\mathbb{E}}_t[...]$ indicates an empirical average over a finite batch of samples, stochastic policy $\pi_{\theta}$ predicts the probability of taking action $a_t$ at state $s_t$, and $\hat{A}_t$ is an estimator of the advantage function at timestep $t$.  
Suppose each action $a=(a^0,\ldots,a^{N_a-1})$, then the objective after action decoupling becomes: 
\begin{equation}
\footnotesize
\max_{\theta} \sum_{i=0}^{N_a - 1} \hat{\mathbb{E}}_{s, a\sim \pi_{\theta_{\text{old}}}}\bigg[ \frac{\pi_{\theta}(a_t^{(i)}|s_t)}{\pi_{\theta_{\text{old}}}(a_t^{(i)}|s_t)}
\hat{A}_t \bigg].
\end{equation}
This decoupled objective brings two advantages. First, it simplifies the policy structure. Specifically, the policy network can be defined without considering the inter-correlations as this dependency can be post-processed. 
Second, it increases the diversity of actions. As each component has its independent own channel of value output, the actions can be significantly diversified, thus inducing more explorations during training. 
Furthermore, to force diversity of exploration, we randomize the positions
%and properties (health, mana, gold, etc.) 
of both agents  during training at the beginning of the game.

%Specifically, at some state $s$, a part of skills, i.e., corresponding to some elements in one action, are invalid because of the cooling-down of skills or blocking of obstacles. The AI server first detects whether a skill is accessible and then encode the accessibility information as a mask including 0 or 1. 
%With action mask, we calculate the dot-product between action masks and $h_{LSTM}$ to predict the final action $a$ and the parameters $p$. The output space is then pruned with action masks which contributes to sampling efficiency.
 
% For example, the direction of a skill always depends on which kind of the skill is predicted. Let $p_t$ denote the corresponding component for each action $a_t$, then the objective can be written as:
% \begin{equation*}
% \max_{\theta} \hat{\mathbb{E}}_{s, a\sim \pi_{\theta_{\text{old}}}}\bigg[ \frac{\pi_{\theta}(a_t|s_t)\pi_{\theta}(p_t|s_t, a_t)}{\pi_{\theta_{\text{old}}}(a_t|s_t)\pi_{\theta_{\text{old}}}(p_t|s_t, a_t)} \hat{A}_t \bigg]
% \end{equation*}

%We instead decouple the inter-correlations in one action by treating each component as an independent channel, and optimize the following objective:
% \begin{equation*}
% \max_{\theta} \sum_{i=0}^{N_a - 1} \hat{\mathbb{E}}_{s, a\sim \pi_{\theta_{\text{old}}}}\bigg[ \bigg( \frac{\pi_{\theta}(a_t^{(i)}|s_t)}{\pi_{\theta_{\text{old}}}(a_t^{(i)}|s_t)} +   \frac{\pi_{\theta}(p_t^{(i)}|s_t)}{\pi_{\theta_{\text{old}}}(p_t^{(i)}|s_t)} \bigg)
% \hat{A}_t \bigg]
% \end{equation*}

However, the action decoupling further increases the complexity of policy training, while it is originally very high due to the vast action and state spaces in MOBA 1v1 games. To improve the training efficiency,  an \textbf{action mask} is proposed to incorporate the correlations between action elements at the final output layers of the policy based on prior knowledge of experienced human player, which helps prune the exploration of RL.
Specifically, our action mask helps eliminate several unreasonable aspects:
1) physically forbidden areas on map, e.g., suppose the predicted action is to move towards a direction, which cannot be performed as that direction is occupied by obstacles in the map; 
2) skill or attack availability, e.g., the predicted action to release a skill within Cool Down time shall be eliminated; 
3) being controlled by enemy hero skill or equipment effects; 
4) hero-/item-specific restrictions. 

\paragraph{Dual-clip PPO}
%Note that the proposed framework is highly distributed with a large number of asynchronously updated learners and actors in an off-policy manner. 
%In this context, the samples collected for training are likely to be subject to dramatically varied distributions that deviate from the current policy. 
%The standard PPO~\cite{schulman2017proximal} may fail to work with these samples since it was originally proposed for on-policy training. %, where the samples share similar distributions with current policy. 
%To address this issue, 
Let $r_t(\theta)$ denote the probability ratio $\frac{\pi_{\theta}(a_t|s_t)}{\pi_{\theta_{\text{old}}}(a_t|s_t)}$. 
Because the ratio $r_t(\theta)$ can be extremely large, maximization of the RL objective may lead to an excessively large policy deviation. 
To alleviate this issue, the standard PPO algorithm~\cite{schulman2017proximal} involves a ratio clip as follows:
%modifies the standard RL objective to the following one:
\begin{equation}
\footnotesize
\mathcal{L}^{\text{CLIP}}(\theta) = \hat{\mathbb{E}_{t}}\Big[ \min\Big( r_t(\theta)\hat{A}_t, \text{clip}\big(r_t(\theta), 1 - \epsilon, 1 + \epsilon\big)\hat{A}_t \Big) \Big],
\end{equation}
to penalize extreme changes to the policy.

\begin{figure}
    \centering
    \includegraphics[width=0.9\linewidth]{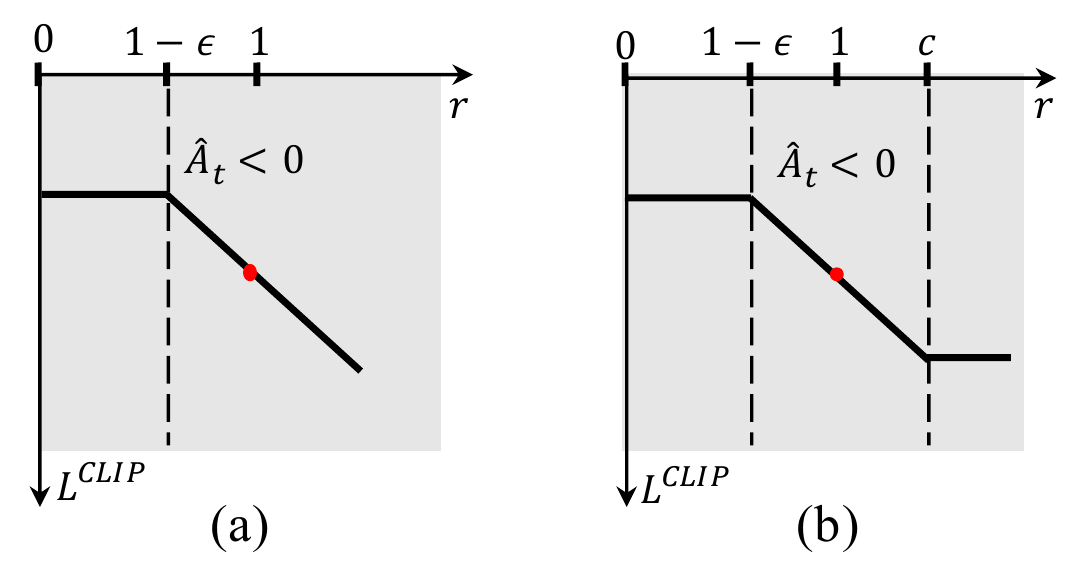}
    \vspace{-2mm}
    \caption{(a) Standard PPO (clip with $\epsilon$); (b) Our Proposed Dual-clip PPO ( clip with $\epsilon$ and $c$ when $\hat{A}_t < 0$)}
    \label{fig:clipping}
\vspace{-3mm}
\end{figure}

However, in large-scale off-policy training environments like our framework, the trajectories are sampled from various sources of policies, which may differ considerably from the current policy $\pi_{\theta}$. 
In such situations, the standard PPO will fail to work with such deviations since it was originally proposed for on-policy~\cite{schulman2017proximal}. 
For example, when $\pi_{\theta}(a_t^{(i)}|s_t) \gg \pi_{\theta_{\text{old}}}(a_t^{(i)}|s_t)$, the ratio $r_t(\theta)$ is a huge number.
When $\hat{A}_t < 0$, such a large ratio $r_t(\theta)$ will introduce a big and unbounded variance since $r_t(\theta)\hat{A}_t \ll 0$. 
As a result, even using the objective of PPO, the new policy deviates significantly from the old policy, which makes it very difficult to insure the policy convergence. 
We thus propose a dual-clipped PPO algorithm to support large-scale distributed training, which further clip the ratio $r_t(\theta)$ with a lower bound of the value $r_t(\theta)\hat{A}_t$, illustrated in Fig. \ref{fig:clipping}. 
When $\hat{A}_t < 0$, the new objective of our dual-clipped PPO is:
\begin{equation}
\begin{split}
\footnotesize
%\mathcal{L}(\theta) = 
\hat{\mathbb{E}_{t}}\Big[\max\Big( \min\Big( r_t(\theta)\hat{A}_t, \text{clip}\big(r_t(\theta), 1 - \epsilon, 1 + \epsilon\big)\hat{A}_t \Big), c\hat{A}_t\Big) \Big]
\end{split}
\end{equation} where $c>1$ is a constant indicating the lower bound.
%Our empirical results show the proposed dual-clipped PPO that the training can not converge without . % (tested on \textit{Honor of Kings}). 

\section{Experiments}

\subsection{System Setup} 
We test our method by using the 1v1 mode in \textit{Honor of Kings}, which is the most popular MOBA game nowadays, and has been actively used as the testbed for recent RL advances~\cite{eisenach2018marginal,wang2018exponentially,jiang2018feedback}. 

Our framework runs over a total number of 600,000 CPU cores encapsulated in Dockers and 1,064 Nvidia GPUs (a mixture of Tesla P40 and V100). 
The volume of our framework allows parallel experiments. 
%We transform agent observations to features. 
We have 1600 vector features 
%(the length varies slightly among heroes as each hero has its own particularity), 
containing observable unit attributions and game information, and 2 channels of image features read from game core (the obstacle channel and the hero position channel).
We mainly use vectors to represent observations as they they are lightweight. 
%The size of observation extracted from game is 1.56 KB. 
%We use 3 layers of FC/ReLu to encode the input vectors. 
We adopt FP16 for data transmission to save bandwidth, and revert to FP32 for training. 
%The memory pool in our framework is implemented by using a lock-free queue structure with size 800,000, to restore the pair of $\{s_i, a_i, r_i, s_{i+1}\}$ sent from AI servers. 
%In each iteration, we randomly select samples from the Memory Pool. 
To train one hero, we use 48 P40 GPU cards and 18,000 CPU cores.
%The batch size is in total 262144, and 
The mini-batch size per GPU card is 4096.
The time step and unit size of the LSTM are 16 and 1024, respectively. 
We train using full rollouts, i.e., one episode ends until the termination of the game, and we use zero-start, i.e., the agent starts the game from Frame 0. 
The training speed is about 80000 samples per second per GPU card. 
We use the game core of \textit{Honor of Kings} directly to execute the game. 
%which could significantly decrease the game frame synchronization time. 
%The training speed 
With the high throughput of our framework, we have experiences collected per day per hero is about 500 years human data in the 1v1 mode of \textit{Honor of Kings}. 
 %We will make our framework and the game core open to the public, to faciliate RL research in complex games. 
%The batches per minute generated are about 340 
%\textcolor{red}{Our framework is able to produce XXX samples per second, xxx . }
%, including Gamecore for running the game and AIServer to implement feature extraction, model prediction and reward computing.

We use Adam optimizer with initial learning rate 0.0001. 
In the dual-clipped PPO, the two clipping hyperparameters $\epsilon$ and $c$ are set as 0.2 and 3, respectively. 
%Since in each state, our agent is trained to maximize the exponentially decayed sum of the future rewards,
The discount factor is set as 0.997. 
For the case of \textit{Honor of Kings}, this discount is valuing future rewards with a half-life of about 46 seconds. 
We use generalized advantage estimation (GAE)~\cite{schulman2015high} for reward calculation, 
and we set $\lambda = 0.95$ in GAE to reduce the variance caused by delayed effects. 

%It contains a fixed pool of opponent models and we require each new model to compete against all of the early models for 64 games. 

%The most straightforward way to evaluate the performance of a game AI is to play against top professional players. Following the evaluation of AlphaGo which played against Fan Hui (top human Go player), we invite the world's best King of Glory solo players to compete against our model. 
To evaluate the trained AI's ability in real world, we deploy the AI model into \textit{Honor of Kings} to play against professional and top-amateur human players. We predict actions via the AI model every 133 ms which is about the response time of top-amateur players. We also compare our AI with baseline methods used in existing works \cite{jiang2018feedback} for playing 1v1 games in \textit{Honor of Kings}, including behavior-tree (implemented as the internal AI in the game), MTCS and its variants. 
We also use Elo rating~\cite{coulom2008whole} for comparing different versions of models, similar to that of AlphaGo \cite{silver2016mastering}. 
%Elo is a method for calculating the relative skill levels of players in zero-sum 1v1 games. 
%A higher Elo score indicates a stronger AI ability.
%which has been widely used in building traditional AI for MOBA; 2) tree-search based RL \cite{jiang2018feedback}, which has been used to build an AI for playing 1v1 games in \textit{Honor of Kings}.   % previously applied on AI for \textit{Honor of Kings}. 

%We study the effectiveness 
%For estimating the ability of trained AI agents, we build a ELO score [\ref{item:elo}] evaluating system. It contains a fixed pool of opponent models and we require each new model to compete against all of the early models for 64 games. 
%After collecting the winning rate, we can calculate ELO score for the new model. A higher ELO score indicates a stronger AI ability. 

%\subsection{Evaluation Metrics} 

%In this experiment, we use two metrics to evaluate the performance of our models, winning rate for comparing the ability of two models and ELO score for comparing a group of models more than two.

%\begin{enumerate}
%    \item \textbf{Winning rate}: The most straightforward metric to evaluate the performance of two models is the winning rate against the other model or the human performance.
%    \item\label{item:elo} \textbf{ELO score}: A method for computing the relative skill ability of players in zero-sum games such as MOBA games or chess, created by Arpad Elo. We build a ELO score evaluating system to evaluate the ability of each models.
%\end{enumerate}

%\subsection{System Performance}

\subsection{Experimental Results}

\begin{table}[t!]
\centering
\scriptsize
\caption{Information of human professional testers. Here ``the league'' refers to the \textit{Honor of Kings} Professional League. }
\vspace{-0.1in}
\begin{tabular}{ll}
\toprule
\textbf{Player / Esports Club}  & \textbf{Info}       \\ \midrule
Main Mage of eStarPro (2019) &    professional, a.k.a, best  Mage  in the league \\ 
Main Marksman of QGhappy (2019) & professional, a.k.a, best Marksman  in the league\\ 
Main Assassin of TS (2019) &    professional, famous top Assassin player  \\ 
Main Warrior of QGhappy (2019)    &   professional, a.k.a best Warrior in the league \\ 
Main Warrior of WE (2019)  &  professional, famous top Warrior player \\ 
\bottomrule
\end{tabular}
\label{table:player_info}
\vspace{-3mm}
\end{table}

\begin{table*}[t!]
\centering
\scriptsize
\caption{Match Statistics of our AI vs. Professional Players on Different Types of Heroes}
\vspace{-0.1in}
\begin{tabular}{cccccc}
\toprule
\textbf{Hero} & \textbf{DiaoChan} & \textbf{DiRenjie} & \textbf{LuNa} & \textbf{HanXin} & \textbf{HuaMulan} \\ \midrule
Hero Type & Mage & Marksman & Warrior+Mage & Assassin & Warrior \\ 
%Score (BO5) & 3:0 (AI:eStarPro.Cat) & 3:0 (AI:QGhappy.Hurt) & 3:0 (AI:QGhappy.Fly) & 3:1 (AI:TS.NuanYang) & 3:0 (AI:WE.762) \\ 
Score (BO5) & 3:0 (AI:Professional)  & 3:0 (AI:Professional)  & 3:0 (AI:Professional)  & 3:1 (AI:Professional) & 3:0 (AI:Professional) \\ 
Kill & 5.0:1.3  & 2.3:0.7 & 2.7:1.0 & 2.5:1.5 & 4.0:1.3 \\ 
Game Length & 6'56'' & 6'23'' & 7'53'' & 6'41'' & 6'48'' \\ 
Gold/min & 852.7:430.6 & 869.3:606.6 & 969.7:724.0 & 954.1:754.2 & 945.2:654.2 \\ 
Exp/min & 900.0:573.0 & 895.3:661.7 &979.0:817.2 & 965.4:802.5 & 921.4:723.1 \\ \bottomrule
\vspace{-4mm}
\end{tabular}
\label{table:match statistics}
\end{table*}

\begin{table}[t!]
\centering
\scriptsize
\caption{Results of AI vs. Various Top Human Players}
\vspace{-0.1in}
\label{table:cj_results}
\begin{tabular}{lllll}
\toprule
\textbf{Hero Name} & \textbf{Hero Type} & \textbf{\#Matches} & \textbf{\#Win}  & \textbf{Rate} \\ \midrule
DiaoChan & Mage & 445 & 445&   100\% \\
DiRenJie & Marksman & 264& 264  &  100\% \\ 
HuaMuLan & Warrior & 256 & 256 &  100\% \\
HanXin & Assassin & 221 & 220  & 99.55\% \\
LuNa & Warrior+Mage & 260 & 260  & 100\% \\
HouYi & Marksman & 79 & 78  & 98.70\% \\
LuBan & Marksman & 354 & 354  & 100\% \\
SunWukong & Assassin & 221 & 219 & 99.09\% \\
\midrule
 & & 2100 & 2096 & 99.81\% \\ 
\bottomrule
\end{tabular}
\end{table}

\subsubsection{Exploring the Upper Limit of Control Ability} We invite 5 top professional \textit{Honor of Kings} human players to play BO5 (best of five) matches against our AI. 
They are all active professional players of famous Esports clubs, and can represent the highest level of hero control ability of human.  
%\texttt{QGhappy.Hurt}, \texttt{WE.762}, \texttt{TS.NuanYang}, \texttt{QGhappy.Fly}, \texttt{eStarPro.Cat} (name in \texttt{CLUB.PLAYER} format). 
The information of these players are briefed in Table \ref{table:player_info}. 

Table \ref{table:match statistics} shows the match results. Note that these professional players play the heroes they are specialized in. 
We see that our AI can defeat professional players on  heroes of various types. 
Take the Mage hero \texttt{DiaoChan} for instance, 
our AI defeats the professional human opponent with score 3:0. 
The human opponent is the current best Mage player in the professional league, and is particularly good at controlling DiaoChan.  
We can see that \texttt{DiaoChan} controlled by AI dominates the game. It achieves 5 kills per game, but gets killed only 1.33 times on average. 
The game lasts for 6 minutes and 56 seconds on average. 
AI also prevails significantly in terms of gold and experience gained in game.

\subsubsection{Evaluating the Robustness of Control Ability}
We further evaluate whether the policies learned by our AI could counter to a diversity of top human players. %, in case that it might be overfitted to professional players. 
% In ChinaJoy 2019, we open our AI in  public matches for four days, from Aug. 2 to Aug. 5, 2019, held in Shanghai, China. 
In ChinaJoy 2019, we held large public matches where the public was allowed to face off against our AI, from Aug. 2 to Aug. 5, 2019, held in Shanghai, China~\footnote{This event was held jointly by Vivo, Qualcomm and Tencent.}. 
We showcased eight AI heroes of different types in total. 
%, containing all types of heroes suitable to play solo games (Warrior, Mage, Marksman, Assassin). 
Human players who defeat AI in 1v1 games will be rewarded with a 600 USD smart phone. 
%Participants must have demonstrated rankings in \textit{Honor of Kings} as the entry condition (top 1\% minimum).  
Participants are required to have demonstrated rankings in \textit{Honor of Kings}. 
% as the entry condition (top 1\% minimum).  
%Before this public event, we 
The statistics results of our public experiment are provided in Table \ref{table:cj_results}. 
Our AI achieves a 99.81\% win rate among 2100 matches, with only 4 games lost. 
Five of the eight heroes achieve a 100\% win rate through hundreds of matches. 
%In these 4 games, AI lose to 
%The China No.1 amateur HanXin player 

\subsubsection{Comparison with Baselines}
\begin{figure}[h!]
    \centering
\includegraphics[width=.95\linewidth]{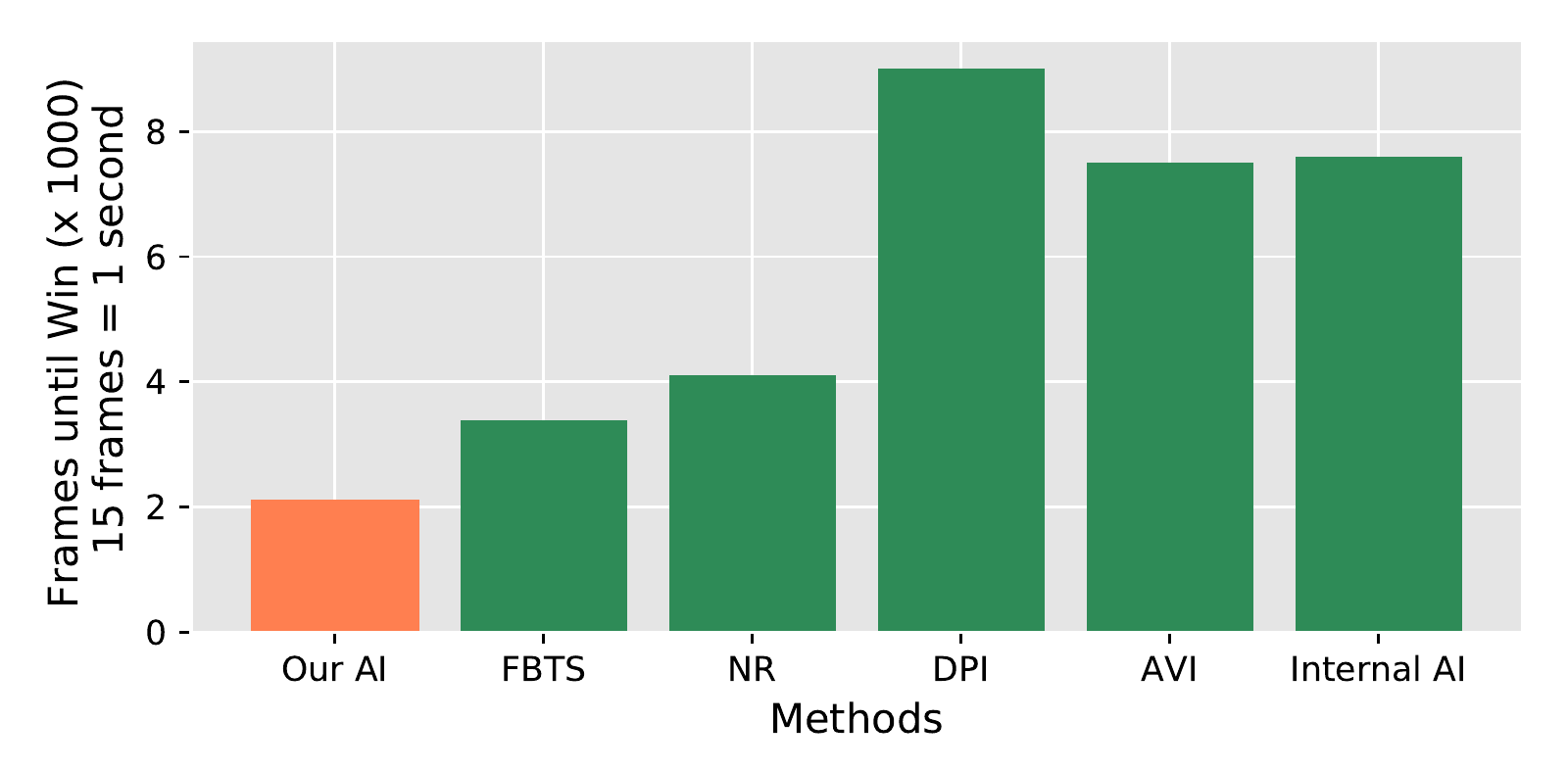}
    \vspace{-2mm}
    \caption{Comparing Averaged Time Length to Defeat the Same Baseline Opponent} \label{fig:comp_jiang}
    \vspace{-4mm}
\end{figure}

We also compare our method with existing methods. 
%First, we compare with the behavior tree based method, which is a classical solution for building AI agents in MOBA. 
%We implement 
As mentioned, a recent work applies monte-carlo tree search (MCTS) based RL to build AI for playing 1v1 games in \textit{Honor of Kings} \cite{jiang2018feedback}. 
Particularly, they built four AI agents with MCTS and its variants, which are named as ``FBTS'', ``NR'', ``DPI'', and ``AVI'', respectively. 
Their evaluation method is to compare which of these agents can defeat the same baseline opponent faster, while the detailed match statistics have not been reported. 
Specifically, their baseline opponent consists of six Marksman heroes that can be defeated by the internal ``DiRenJie'' AI, implemented using behavior-trees.  
Following the same experiment settings, we compare the averaged length of time for our AI to defeat the same baseline opponent. 
%The shorter the time length, the stronger the AI's ability. 
The result is shown in Fig. \ref{fig:comp_jiang}. 
We see that the AI trained from our method significantly outperforms the FBTS, NR, DPI, AVI and the internal AI. 
%, i.e., choosing the six Marksman heroes that can be defeated by the internal ``DiRenJie'' AI. And we calculate the average amount of time 

\subsubsection{Progression During Training}

\begin{figure}[t!]
    \centering
    \includegraphics[width=\linewidth]{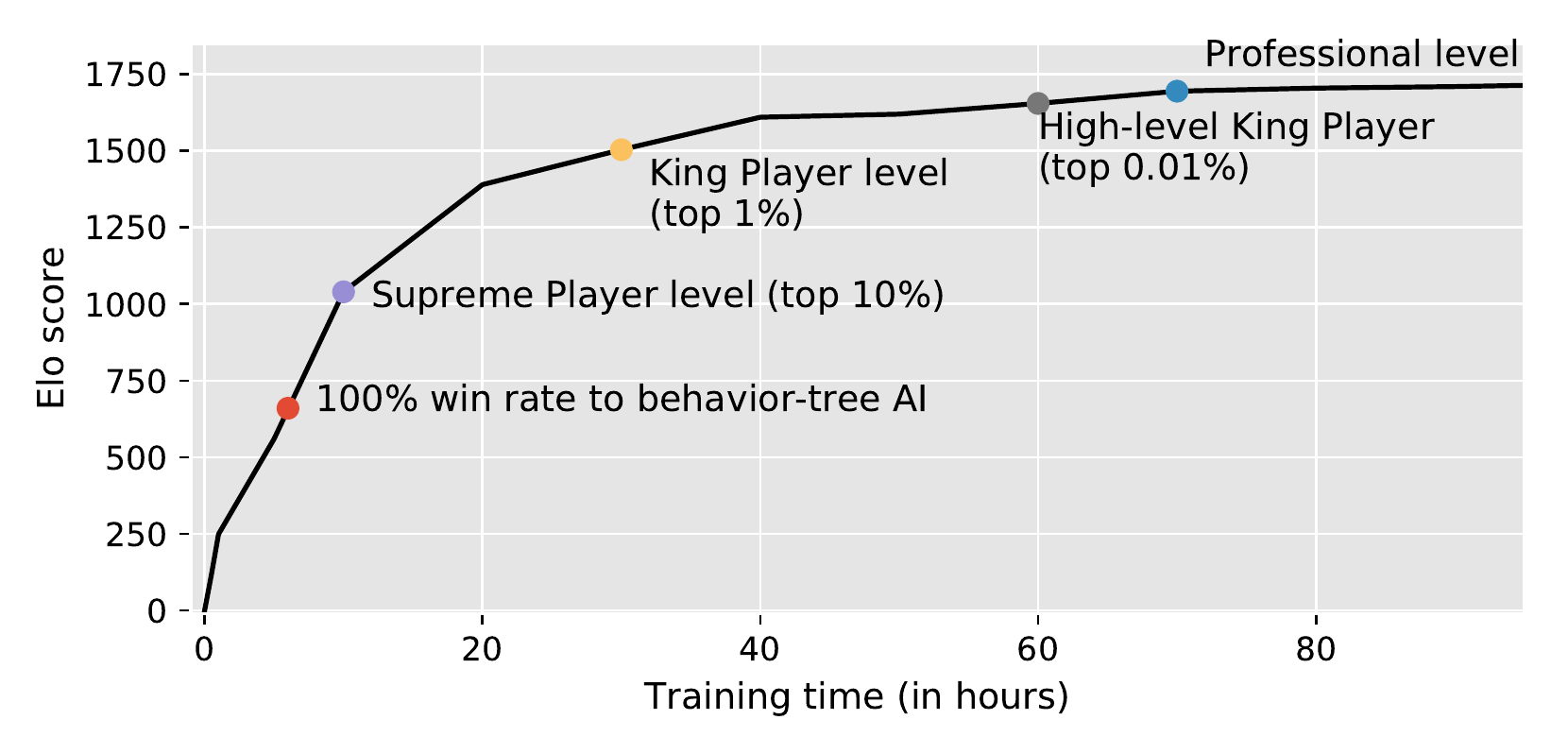}
    \vspace{-0.3in}
    \caption{Elo Change during Training}
        \vspace{-2mm}
    \label{fig:elo_change}
\end{figure}

In Fig. \ref{fig:elo_change}, we show the change of Elo rating during training, using the Marksman hero ``DiRenJie'' as a case. 
We observe that the Elo score grows with the training length and becomes relatively steady after about 80 hours. 
Further, the growth rate of Elo is inversely proportional to training time. 
With about 6 hours training, the AI begins to defeat the internal behavior-tree based AI with 100\% win rate. 
The AI ability rapidly increases to King Player's level (top 1\% human player for \textit{Honor of Kings}) after about 30 hours, and becomes comparable to professionals after 70 hours. 

\subsubsection{Ablation Study}

\begin{table}[t!]
\centering
\scriptsize
\caption{Results of Ablation Experiments} \vspace{-3mm}
\label{tab:ablation}
\begin{tabular}{lcc}
\toprule
\textbf{Item} & \textbf{Win rate vs Base} & \textbf{Time to converge}  \\ \midrule
Base & - & 80 h \\
Base + AM & 50.5\% & \textbf{65 h} \\ 
Base + TA & \textbf{75\%}  &   90 h \\
%Base+AM & 50.5\% &  \textbf{65 h} \\ 
Base + LSTM & 73\% &  100 h  \\ 
%Base+FR & 68\% & 90 h  \\ 
%Base+RIF & \textbf{45\%} & 70 h \\ \midrule
Full version & 90\% & 80 h  \\ \bottomrule  
\vspace{-3mm}
\end{tabular}
\end{table}

We conduct ablation experiments to understand the effect of different components and settings in our method. 

We first analyze three components from the model network, including the \textit{action mask} (AM), \textit{target attention} (TA) and \textit{LSTM} we developed. 
In Table \ref{tab:ablation}, we show the results of a tournament between different ``DiRenJie'' AI versions using the same amount of training resources. 
The Full version refers to the strongest AI we trained, while the Base version means the Full version without the aforementioned three components. 
We see that using \textit{action mask} can largely reduce the training time, while achieving the same AI ability as the Base (win rate 50.5\%). 
We see that the TA and the LSTM are both useful to improve the AI ability.  
%For each AI version, we measure time to converge and the win rate against the Base. 

We also analyze hyperparameter settings. 
%, including \textit{full rollouts} (FR) and \textit{random initial frame} (RIF). 
Particularly, we compare \textit{full rollouts} (FR) with \textit{partial rollouts} (PR) with fixed N frames, and we compare \textit{random initial frame} (RIF) with zero-start (ZS), i.e., selecting random frame of the whole game as the start of the Markov decision process (RIF) versus starting from the beginning of the game (ZS). We find that: 1) FR improves the AI's ability to a large margin, with which the win rate increases to 70\%$\sim$80\% when compared to PR with 1000, 2000 and 3000 frames; 2) RIF can speed up the convergence by 15\% but at a cost of slightly lower AI ability (win rate 40\% when compared to ZS). 

\section{Conclusion and Future Work}
%We study the problem of complex action control in a competitive multi-agent setting through MOBA 1v1 games. 
%We study the problem of complex action control in MOBA 1v1 games. 
%in MOBA 1v1 games. 
In this paper, we present a deep reinforcement learning (DRL) approach to handle the complex action control of agents in MOBA 1v1 games, from the perspectives of both system and algorithm. 
We propose a scalable and off-policy DRL system architecture for massive episode exploration. 
We propose an actor-critic multi-label neural network, containing several strategies for modeling MOBA combats, and a dual-clipped version of the PPO algorithm for ensuring convergence. 
The resulting AI from our framework can defeat top professional esports players in MOBA 1v1 games, tested on the popular MOBA game \textit{Honor of Kings}. 

As a next step, we will make our framework and algorithm open source, and the game core of \textit{Honor of Kings} accessible to the community to facilitate further research on complex games; and we will also provide part of our computing resources via virtual cloud for public use \footnote{By Nov. 21, 2019, the Beta version of our framework is open to 4 universities in China for user feedback. }. 

\section{Appendix}

\subsection{MOBA 1v1 Games}

Real-time Strategy (RTS) games are considered as a grand challenge for AI research \cite{alphastarblog}. 
%RTS can be further categorized into generaly RTS games, e.g., StarCraft, and Multiplayer Online Battle Arena (MOBA) games, e.g., Honor of Kings. 
MOBA is one type of RTS games, and is the most played game type \cite{mora2018moba}. 
Popular MOBA games include \textit{Dota}, \textit{Honor of Kings}, \textit{League of Legends}, etc. 

MOBA 1v1 mode is a pure arena for competing one's level of action control.
%In this mode, a player control 
The formal 1v1 matches, for fairness, are mirror games, i.e., two players pick the same hero and control their own hero individually to fight against each other. %, thereby enabling fair competition. 
When the game begins, each player sets out from the base, gains gold and experience by killing or destroying other game units (e.g., enemy heroes, creeps, turrets). 
The goal is to destroy the opponent's turrets and base while protecting own turrets and base.
Briefly, creeps are a small group of computer-controlled creatures periodically travel along the predefined lane to attack the opponent. Turrets, i.e., defensive buildings, are designed to attack any creeps or heroes moving into their sight area. 
Heroes are player-controlled units which can move around and have abilities to release various attacks and healing skills. 

\begin{figure}[t!]
    \centering
    \includegraphics[width=0.95\linewidth]{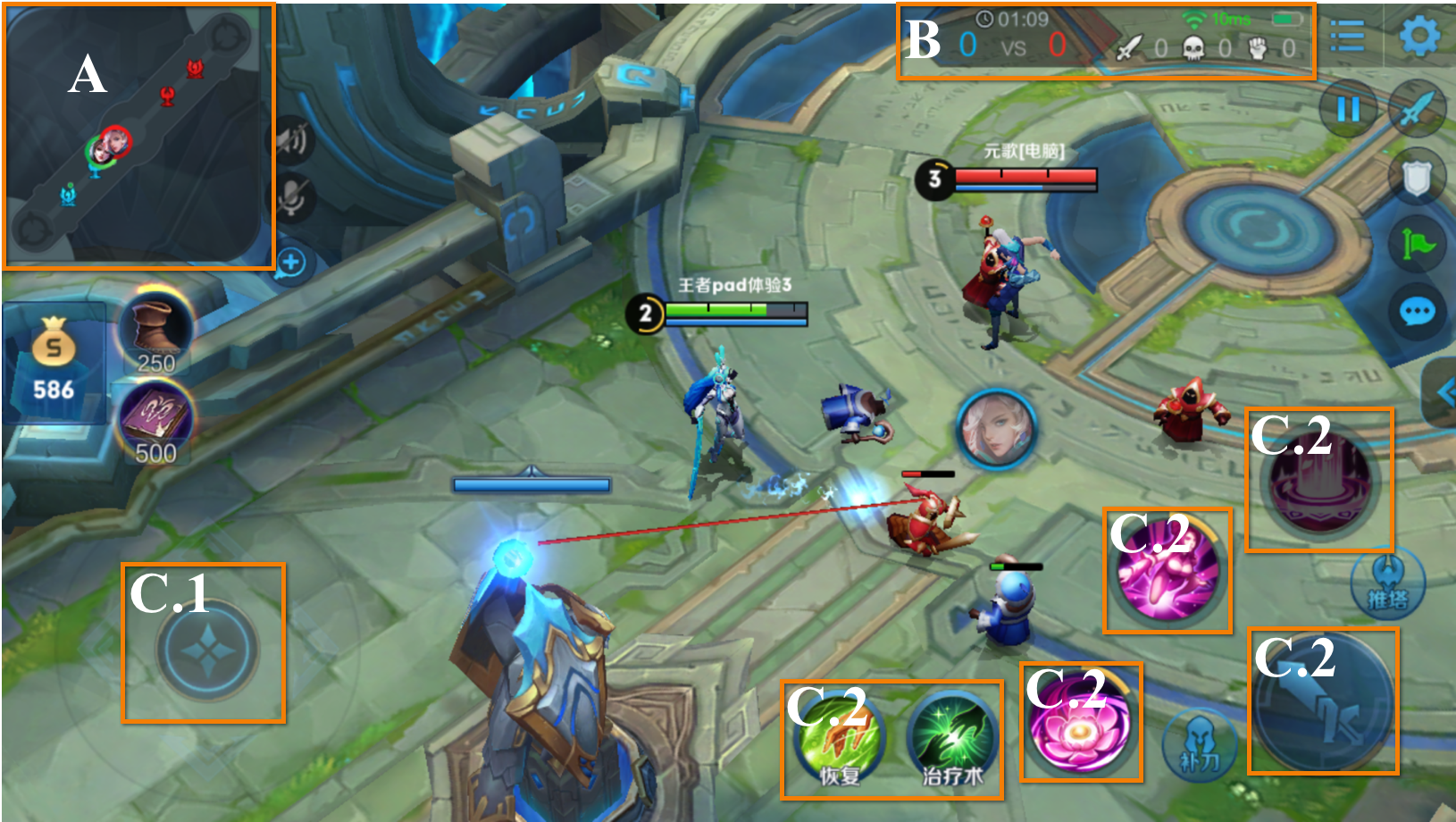}
    \caption{Game UI of \textit{Honor of Kings} 1v1. In the main screen, there are four sub-parts: mini-map (A) on the top-left, dashboard (B) on the top-right, movement controller (C.1) on the bottom-left, and ability controller (C.2) on the bottom-right, as highlighted in each box. }
    \label{fig:interface}
    \vspace{-3mm}
\end{figure}

Each MOBA hero requires very complicated control mechanism, known as micro-management in esports. 
%By comparison, for general RTS games like StarCraft, the player has to control many more game units than that of MOBA, but the control mechanism of the player's hero is much simpler. 
All MOBA games have various types of heroes. 
For example, in \textit{Honor of Kings}, there are six types of heroes, including Mage, Assassin, Warrior, Marksman, Tank, Support. 
Different hero types have different playing method. Generally, Tank and Support are durable heroes and are mainly for defense, thus not suitable for MOBA 1v1 solo games. 
The rest four types are generally selected for MOBA 1v1 esports. 
%, each of them takes advantages on several aspects.
In Fig. \ref{fig:interface}, we show the UI of the MOBA game \textit{Honor of Kings} as an example. 

\begin{table}[t!]
\centering
\scriptsize
\caption{Reward Design} 
%\vspace{-3mm}
\label{tab:reward}
\begin{tabular}{llll}
\toprule
\textbf{Reward} & \textbf{Weight} & \textbf{Type} & \textbf{Description}  \\ \midrule
hp\_point & 2.0  & dense & the health point of hero \\
tower\_hp\_point & 10.0 & sparse & the health point of turrets and base\\ 
money (gold) & 0.008 & dense & the gold gained\\
ep\_rate & 0.8 & dense & the rate of mana  \\ 
death  & -1.0 & sparse & being killed \\ 
kill & -0.5 & sparse & kill an enemy hero\\
exp & 0.008 & dense & the experience gained \\ 
last\_hit & 0.5 & sparse & last hitting to enemy units \\ \bottomrule
\end{tabular}
\end{table}

\begin{figure}[t!]
    \centering
    \includegraphics[width=0.95\linewidth]{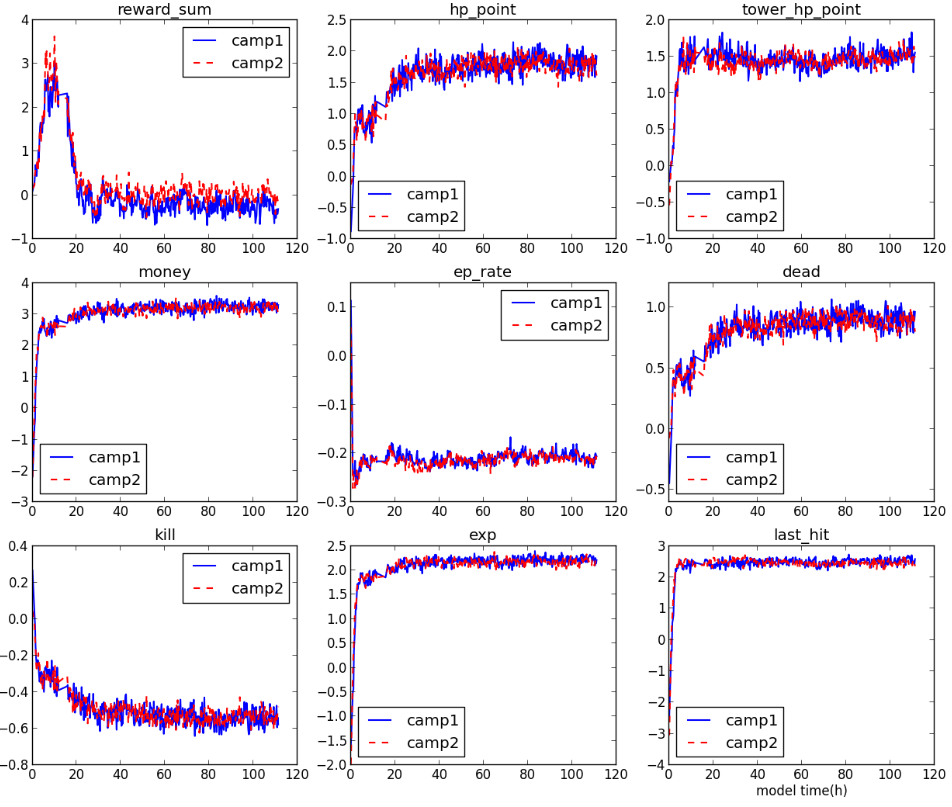} 
    \caption{A case of reward change during training; the x-axis is the training time in hours, the y-axis is reward; camp1 and camp2 refer to the two camps (teams) in MOBA games. }
    \label{fig:reward_curve}
    \vspace{-0.1in}
\end{figure}

\subsection{Reward}
All the trained 1v1 heroes use the same reward, shown in Table \ref{tab:reward}. 
The reward design is inspired by OpenAI Five's Dota reward \cite{OpenAI_dota}. 
The reward is zero-sum, i.e., one's mean reward is subtracted from that of the opponent. 
Our framework allows on-the-fly reward analysis during training. A case of the \texttt{DiaoChan} hero is shown in Fig. \ref{fig:reward_curve}. 

\bibliographystyle{aaai}
\bibliography{AAAI-YeD.1123}
\end{document}